\documentclass[10pt,twocolumn,letterpaper]{article}

\usepackage{iccv}
\usepackage{times}
\usepackage{epsfig}
\usepackage{graphicx}
\usepackage{amsmath}
\usepackage{amssymb}

\usepackage[breaklinks=true,bookmarks=false]{hyperref}

\iccvfinalcopy % *** Uncomment this line for the final submission

\def\iccvPaperID{****} % *** Enter the ICCV Paper ID here

\ificcvfinal\fi

\newcommand{\draftyamamoto}[1]{{\color{black} #1}} %change points after proofreading

\makeatletter
\newcommand{\printfnsymbol}[1]{%
  \textsuperscript{\@fnsymbol{#1}}%
}
\makeatother
\begin{document}

%%%%%%%%% TITLE
\title{Multi-modal Affect Analysis using standardized data within subjects in the Wild }

%\author{Sachihiro Youoku, Junya Saito, Takahisa Yamamoto, Akiyoshi Uchida, Xiaoyu Mi, \\
\author{Sachihiro Youoku\thanks{equal contribution}, Takahisa Yamamoto\printfnsymbol{1}, Junya Saito, Akiyoshi Uchida, Xiaoyu Mi, \\
Osafumi Nakayama, Kentaro Murase \\
Advanced Converging Technologies Laboratories, Fujitsu Ltd.\\
%{\tt\small \{youoku, \}@i1.org}
\and
Ziqiang Shi, Liu Liu, Zhongling Liu\\
Fujitsu R\&D Center Co.,LTD\\
}

\maketitle
\ificcvfinal\thispagestyle{empty}\fi

%%%%%%%%% ABSTRACT
\begin{abstract}
   Human affective recognition is an important factor in human-computer interaction. However, the method development with in-the-wild data is not yet accurate enough for practical usage. In this paper, we introduce the affective recognition method focusing on facial expression (EXP) \draftyamamoto{and valence-arousal calculation} that was submitted to the Affective Behavior Analysis in-the-wild (ABAW) 2021 Contest. 
   When annotating facial expressions from a video, we thought that it would be judged not only from the features common to all people, but also from the relative changes in the time series of individuals. Therefore, after learning the common features for each frame, we constructed a facial expression estimation model \draftyamamoto{and valence-arousal model} using time-series data after combining the common features and the standardized features for each video. Furthermore, the above features were learned using multi-modal data such as image features, AU, Head pose, and Gaze.
   %In the validation set, our model achieved a facial expression score of 0.546 \draftyamamoto{and valence-arousal score of xxxx and xxxx}. These verification results reveal that our proposed framework can improve estimation accuracy and robustness effectively.
   In the validation set, our model achieved a facial expression score of 0.546. These verification results reveal that our proposed framework can improve estimation accuracy and robustness effectively.
\end{abstract}

%%%%%%%%% BODY TEXT
\section{\large{I}\normalsize{NTRODUCTION}}

\begin{figure*}
\begin{center}
\includegraphics[width=16cm]{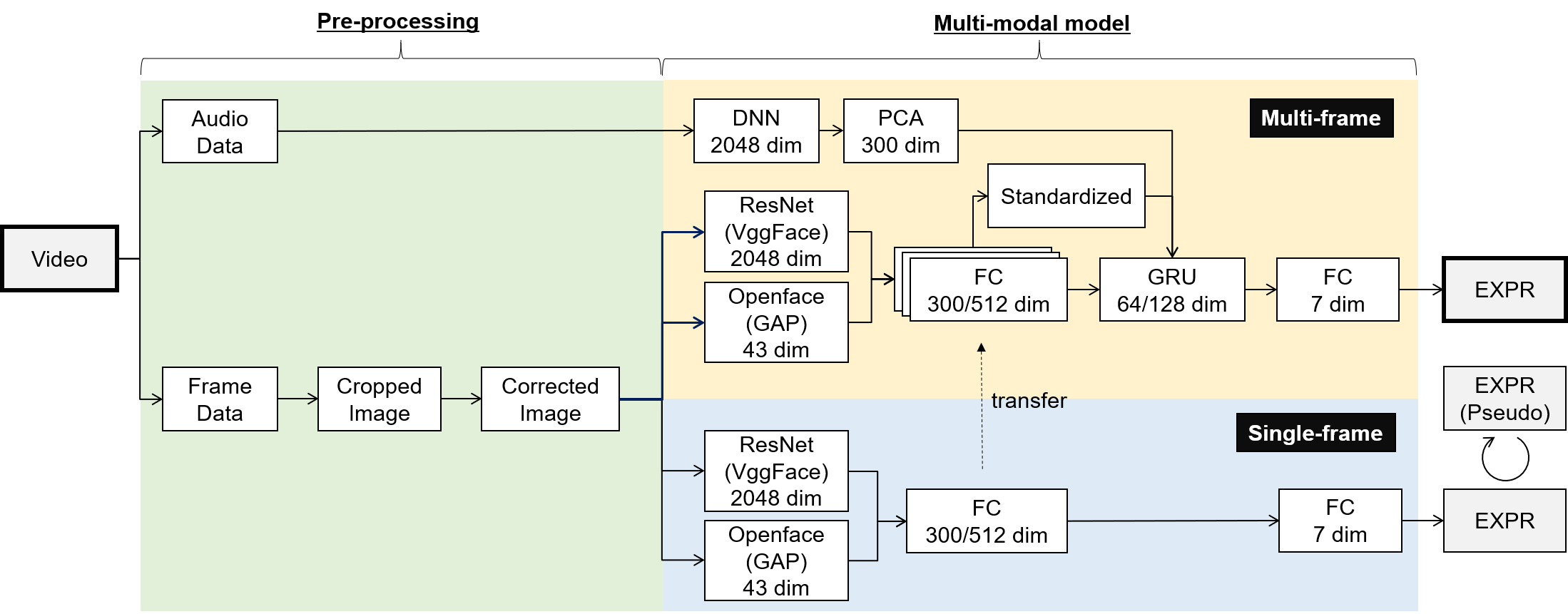}
\end{center}
   \caption{Example of a short caption, which should be centered.}
\label{fig:over_view}
\end{figure*}

Recognizing human affect is becoming a crucial part of human-computer interactive systems. It is expected to contribute to a wide range of fields such as remote healthcare, learning, driver state monitoring, and so on. Many methods to express human mental state have been studied, of which “categorical emotion classification” is one of the most commonly used methods.For the emotional category, the famous six basic emotional expressions\cite{emo1}\cite{emo2} proposed by Ekman and Friesen are popular. Ekman et al. classify emotions as "anger, disgust, fear, happiness, sadness, surprise". 
Recently, D. Kollias has provided a large scale in-the-wild dataset, Aff-Wild2\cite{aff1}\cite{cmp04}. Aff-wild2 is an extended version of Aff-wild\cite{cmp06}\cite{cmp07}. this dataset has used actual videos including a wide range of content (different age, ethnicity, lighting conditions, location, image quality, etc.) collected from YouTube. And multiple labels such as 7 emotion classifications (6 basic emotion expressions + Neutral), Valence-Arousal, Action-unit (based on Facial action coding system (FACS)\cite{au1} have been annotated to the video.
Many complex analyzes of AU, emotion, and VA using this data set have been performed\cite{cmp02}\cite{cmp03}\cite{cmp05}., and large-scale competitions using this data set have also been held\cite{cmp00}\cite{cmp01}.
In this paper, we propose to estimate facial expression using a multi-modal model that leaned common time-series features and  standardized time-series features within subjects. Figure \ref{fig:over_view} shows the framework of the multi-modal model.
In the pre-processing, when video data or image data is received, the face part is detected and cut out, and the color tone is corrected. Facial features are then extracted using a pre-trained model.
Multiple modality features such as action units, head poses, gaze, posture, and ResNet50[\cite{res1}] features are extracted.
In addition, intermediate features that combine these multi-modal features are generated frame by frame. Facial expressions are predicted by inputting the time series values of these intermediate features and the standardized intermediate features of each subject into the GRU\cite{gru01} model.

%%%%%%%%%%%%%%%%%%%%%%%%%%%%%%%%%%%%%%%%%%%%%%%%%%%%%%%%%%%%%%%%%%%%%%%%%%%%%%%%

\section{\large{R}\normalsize{ELATED} \large{W}\normalsize{ORK}}

When dealing with in-the-wild data, the problem is that the color tones of the images are different. A. W. Yip et al.\cite{clr01} compared the accuracy of face recognition between color images and gray-scale images, and found that there was almost no difference in accuracy at a certain high resolution.It also shows that if a pseudo-color image with adjusted color tones is refined from a gray-scale image, the accuracy will be equal to or higher than that of a color image even at low resolution.
In emotion estimation, it has been shown that the estimation accuracy is improved by extracting facial features using RESNET pre-trained with the VggFace2 dataset\cite{abaw1}\cite{abaw2}.
It is also suggested that the accuracy of emotion estimation can be improved by learning with multi-modal information including audio as well as video\cite{abaw2}\cite{abaw3}.
In addition, Saito et al focused on the change in facial expression for each subject and presented a method for estimating the action unit with high accuracy by learning the relative change in facial expression within the same subject as a features\cite{abaw4}.
Also, D-H. Lee shows that estimating unlabeled data using a model trained with labeled data, and retraining that estimated value as a pseudo-label improves estimation accuracy\cite{pseudo01}.
Saito et al. has Improved the estimation accuracy by generating a model based on the hypothesis that the annotator makes a relative judgment from the change in the facial condition of the same subject regarding the judgment of the action unit\cite{abaw4}.

%With regard to the impact of various recognition tasks, D. Kollias et al. have shown that it can improve the performance comparing to each single task model to combine the task models of action-unit detection, emotional classification, and valence-arousal estimation into a fused model\cite{c14}\cite{c15}.

%%%%%%%%%%%%%%%%%%%%%%%%%%%%%%%%%%%%%%%%%%%%%%%%%%%%%%%%%%%%%%%%%%%%%%%%%%%%%%%%
\section{\large{M}\normalsize{ETHODOLOGY}}

In this section, we introduce our method for facial expression analysis. Figure \ref{fig:pre_processing} shows an overview of the overall pipeline. First, video stream and audio stream are extracted from the video. These streams are pre-processed individually.
The video stream first crops the facial image frame by frame, generates single-frame features, and then trains the facial expression estimation model. After that, the weights of the model trained in a single frame are used to generate time-series features and build the final facial expression estimation model.

\subsection{Visual Data Pre-processing}

\begin{figure}[t]
  \centering
  \includegraphics[width=8.2cm]{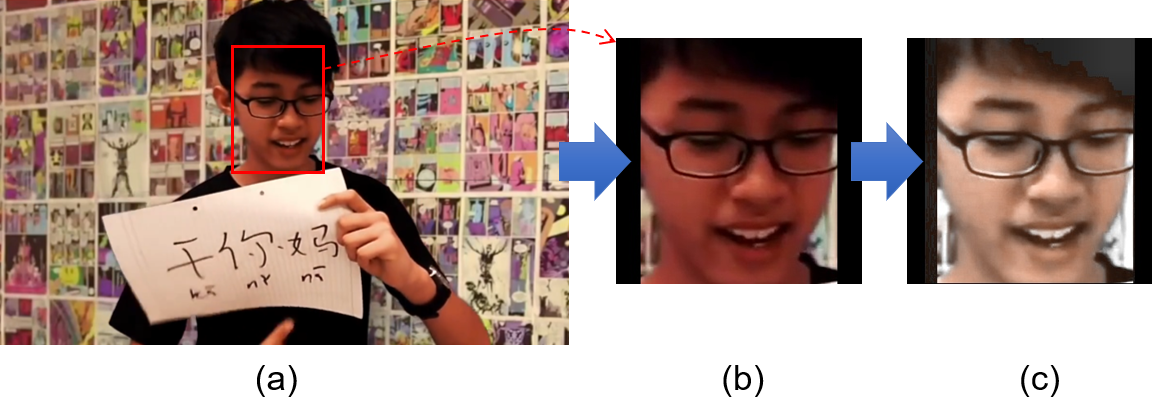}
  \caption{Pre-processing (a) original video frame, (b) cropped face image using MTCNN, (c) corrected face image by i) fixed hue ii) offset saturation, iii)offset value and applied limited adaptive histogram equalization.}
  \label{fig:pre_processing}
\end{figure}

The sequence of Visual Data pre-processing is shown in Figure \ref{fig:pre_processing}.

As a first step, we cut out a facial image from all frames using MTCNN [****]. The Aff-wild2 data also includes a video showing multiple people. We used the coordinates of the MTCNN's bounding box and the heuristic center of gravity of the main subject  to cut out a facial image of the main subject. In addition, the results were cleaned manually.

Next, the color tone of the image was corrected.
In-the-wild data shows variations in skin brightness and color due to different lighting conditions.
Since this variation may cause noise in facial expression estimation, the face image was converted to HSV and corrected as follows.
\begin{itemize}
  \setlength{\itemsep}{0.5mm} 
  \setlength{\parskip}{0.5mm} 
  \item H: Fixed to 14
  \item S: Offset to the average value
  \item V: Offset to the mean, then apply restricted adaptive histogram equalization
\end{itemize}

\subsection{Audio Data Pre-processing}

First, in the pre-processing of audio data, audio was extracted from the moving image.
After that, while shifting the audio data by about one frame, the audio data for the past one second was cut out and the audio features were generated using the DNN described later.

\subsection{Single Frame Model}

\begin{figure}[t]
  \centering
  \includegraphics[width=8.2cm]{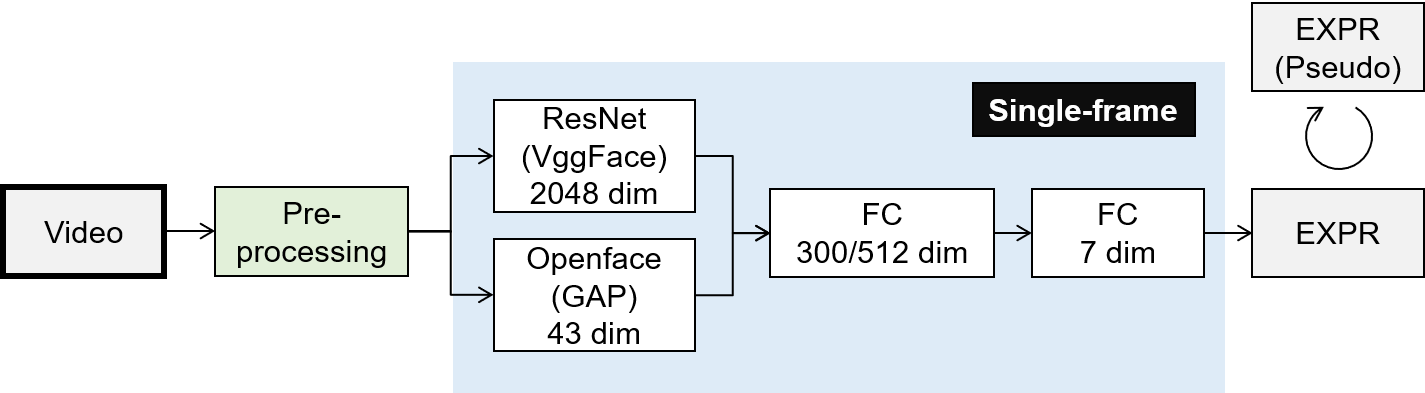}
  \caption{Structure of single-frame model.}
  \label{fig:single_model}
\end{figure}

A single-frame model to estimate facial expression was constructed using the pre-processed images. The sequence is shown in Figure \ref{fig:single_model}.
First, we generated two types of features from the image data. The first used 2048-dimensional features output from a ResNet 50 model pre-trained with VggFace 2. The second were 43-dimensional GAP features, Gaze (2D direction), AU (17D intensity, 18D occurrence), and head pose (3D position, 3D direction), output from openface\cite{of01}.
Next, we learned a single-frame model that estimates facial expressions by combining these two features with the FC layer. The FC layer has two layers, the first layer outputs 300 or 512 dimensions as an intermediate layer, and the second layer outputs seven types of facial expression classifications as the final layer.
Furthermore, using this model, the facial expressions of unlabeled data were estimated and used as pseudo-labels. the single-frame model was generated by retraining with the selected pseudo-labeled and labeled data finally. Since the Aff-wild2 dataset contains not only facial expressions but also valence-arousal labels, the data used for retraining is selected using the following criteria based on Russell circumplex model of affect\cite{circum01}.

\begin{itemize}
  \setlength{\itemsep}{0.5mm} 
  \setlength{\parskip}{0.5mm} 
  \item Data without facial expression label but with valence-arousal label
  \item Pseudo label is 0 and \textbar valence\textbar \textless0.5, \textbar Awakening\textbar \textless0.5
  \item Pseudo label is 1 and valence \textless 0, arousal \textgreater 0
  \item Pseudo label is 2 and valence \textless 0, arousal \textgreater 0
  \item Pseudo label is 3 and valence \textless 0, arousal \textgreater 0
  \item Pseudo label is 4 and valence \textgreater 0, arousal \textgreater 0
  \item Pseudo label is 5 and valence \textless 0, arousal \textless 0
  \item Pseudo label is 6 and awakening \textgreater 0
\end{itemize}

%added by yamamoto%
\draftyamamoto{
\subsubsection{Valence-Arousal}
In this section, we describe the method to calculate valence and arousal.
Our method utilizes three different feature vectors: the one is extracted from cropped images by using ResNet101, the second is made from audio data,
and the third is created from head pose estimation \cite{ref:zhou2020whenet}.
The dimension of image feature and audio feature is reduced to 300 by principal component analysis (PCA), respectively.
After normalizing all three feature vectors, they are concatenated.
As a regression analysis, we have a lot of methods, including CatBoost \cite{ref:emotion_prokhorenkova2018catboost}, SVR, and so forth.
Here, CatBoost is gradient boosted decision tree.
We consider SVR and CatBoost regression analysis methods and evaluate them by using validation data.
}
%%%%%%%%%%%%%%%%%%%%%%%%%%%%%%%%%%%%%%%%%%%

\subsection{Multi Frame Model}

\begin{figure}[t]
  \centering
  \includegraphics[width=8.2cm]{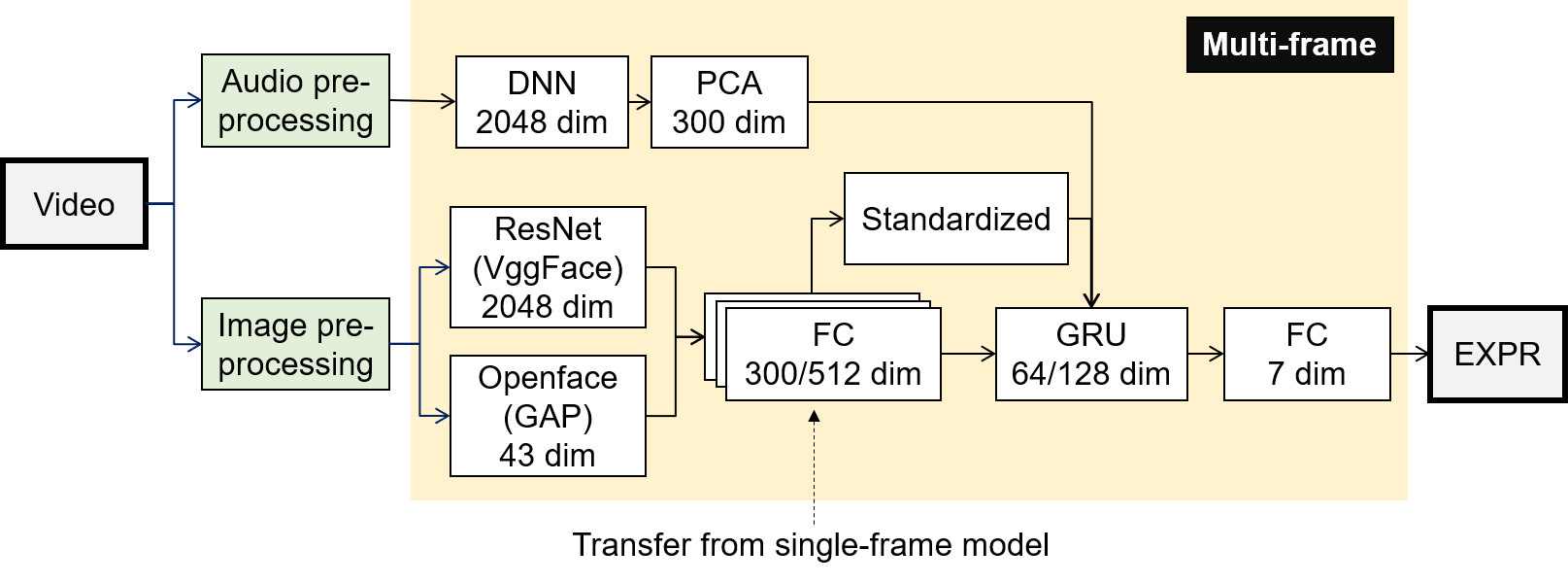}
  \caption{Structure of multi-frame model.}
  \label{fig:multi_model}
\end{figure}

A multi-frame model was built to estimate facial expressions using preprocessed images. The sequence is shown in Figure \ref{fig:multi_model}.
There are two streams, an audio stream and a video stream.
In the video stream, the features using ResNet50 and the features using openface were output from the image as in the case of a single frame. Next, the single frame model was transferred, and the 300 or 512-dimensional features of the intermediate layer were output for each frame.
%In the audio stream, the 2048-dimensional features using DNN were output from the audio data as in the case of a single frame. Next, 300-dimensional features are extracted using PCA. 
\draftyamamoto{To extract deep learning-based audio features, we use a convolutional neural network trained on the large-scale AudioSet containing 5000 hours of audio with 527 sound classes proposed in \cite{ref:kong_audio_ex}. We use the 2048-dimensional output of the second to the last layer as features, so that the four frames of audio corresponding to an image have a total of 8192-dimensional features. To effectively use the features of audio, we have done 
dimension reduction with PCA. To make PCA possible, we sampled 1\% 
of the original audio feature samples, formed a 26274 x 8192 matrix, and perform the PCA with 300 components on it. When the model is obtained, we use it to perform PCA on all the original 8192-dimensional audio features to obtain 300-dimensional deep learning features.
}

Then, image features and audio features are combined, and A multi-frame model was generated by learning multiple frame data that combines the intermediate features and the intermediate features standardized for each subject by GRU.
The data input to the GRU is two-dimensional, which is the feature multiplied by the number of frames, and the N seconds of data was used in the L frame step. Since the video is 30fps, the final data size is calculated as below equation:

\begin{equation}
\label{eq:size}
 datasize = (2 \times (dim_{audio} + dim_{image})) \times \frac{N \times 30}{L} 
\end{equation}
%equation
%gather
$dim\mathrm{_{audio}}$: dimensions of audio features (300)\\
$dim\mathrm{_{image}}$: dimensions of image features (300 or 512)

%%%%%%%%%%%%%%%%%%%%%%%%%%%%%%%%%%%%%%%%%%%%%%%%%%%%%%%%%%%%%%%%%%%%%%%%%%%%%%%%
\section{\large{E}\normalsize{XPERIMENTS}}

\subsection{Implementation and Setup}

[Dataset]

We used the Aff-wild2 dataset\cite{cmp06}\cite{cmp07}. This contains 548 videos, and multiple annotations (Expression, Valence-Arousal, etc.) are added in frame units. This is currently the largest audio-visually annotated in-the-wild dataset. In this challenge, the following Training subjects, Validation subjects, and Test subjects data were provided from the data annotated with Expression.
\begin{itemize}
  \item Expression: 253, 70, 223 subjects in the training, validation, test
\end{itemize}
We also used the Expression in-the-Wild (ExpW) dataset for expression data augmentation.
The ExpW dataset is for facial expression recognition and contains 91,793 faces manually labeled with facial expressions. Like Affwild2, each face image is labeled with seven basic expression categories: "anger", "disgust", "fear", "happiness", "sadness", "surprise", and "neutral". 
\\

[Evaluation Metric]

For Challenge-Track 2: 7 Basic Expression Classification, ABAW Challenge used the accuracy and F1 score, and the score of track 2 is calculated as below equation:
\begin{equation}
\label{eq:exp}
 Score_{expression} = 0.67 * F_1 + 0.33 * Accuracy
\end{equation}

\draftyamamoto{For Challenge-Track 1: Valence-Arousal estimation, the Concordance Correlation Coefficient (CCC) is used for judging the performance of our model as described in this competition guideline.
}
CCC is calculated as below equation:
\begin{equation}
\label{eq:va}
 Score_{ccc. valence/arousal} = \frac{2s_{xy}}{s^2_x + s^2_y + (\bar{x} - \bar{y})^2} 
\end{equation}

[Implementation]

\begin{figure}[t]
  \centering
  \includegraphics[width=8.2cm]{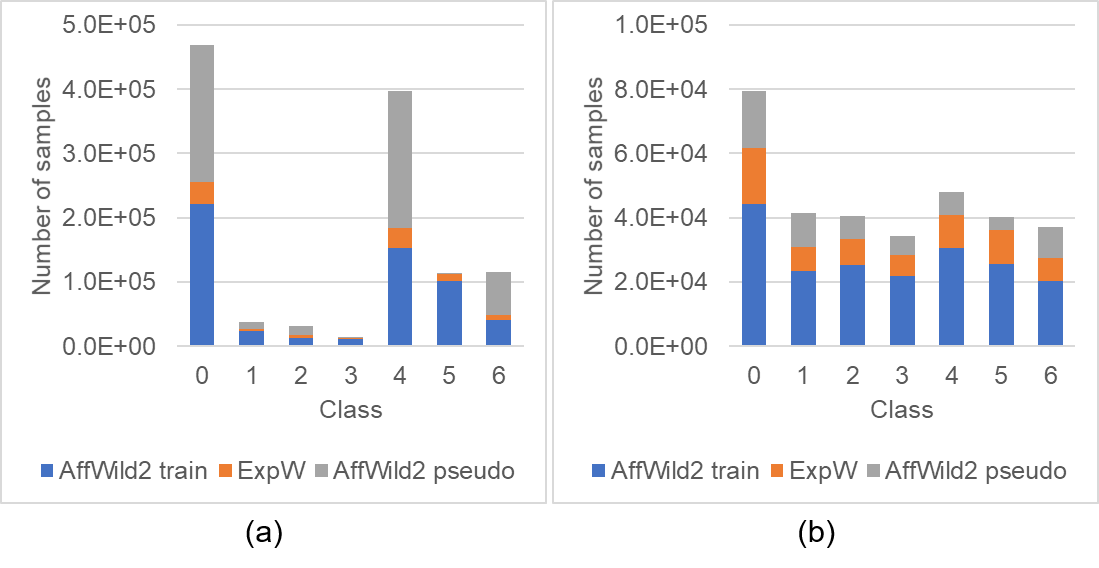}
  \caption{Number of data frames used for training. (a)before balancing, (b)after balancing}
  \label{fig:balance}
\end{figure}

Our framework was implemented by Jupyter Labs. First, I used mtcnn-opencv to cut out the face image. mtcnn-opencv is a library for cutting out the face area using MTCNN\cite{mtcnn01}. The width and height of the image clip is set to 300 pixels. After that, the data was manually cleaned, and finally, the data in the tray of 563,795 and the verification data of 243,006 were used. For pseudo-label, after the above manual data cleaning, selection was performed using Valence-Arousal, and 520,190 data were used. Since the above data is imbalanced, we performed data balancing and finally used the number of frames that showed in figure \ref{fig:balance} for training.
Next, keras-vggface and Openface 2.2.0\cite{of01} were used to extract RESNET-based image features and GAP features, respectively.
We train our GRU model on Aff-Wild2 with the following parameters:\\
\space 1. Length of time N = 2 seconds, step L = 6 frames.\\
\space 2. Length of time N = 3 seconds, step L = 6 frames.\\
In addition, we performed frame-missing interpolation for verification. In the video, there are frames that cannot identify the face, such as "shaking the face" and "covering the face with hands".
Therefore, when there was a frame in which the face could not be identified, the data for the past 30 frames was read and linearly interpolated.

%%%%%%%%%%%%%%%%%%%%%%%%%%%%%%%%%%%%%%%%%%%%%%%%%%%%%%%%%%%%%%%%%%%%%%%%%%%%%%%%
\section{\large{R}\normalsize{ESULTS} AND \large{D}\normalsize{ISCUSSIONS}}

Table \ref{table:validation1} shows the comparison results on the validation set between models trained using our model. The bottom of the table is \textit{Multi-modal-std4} with the addition of overall standardization.
The Expression Score is the result calculated based on Eq. (\ref{eq:exp}). As a result of the validation, it was confirmed that the score of our method is higher than that baseline. In addition, our best score is the same as the score of the first place in the competition last year. However, since the label is different from last year, it is a reference value.
The Score was significantly improved in the model using multi-modal data and standardized data within subjects. We think that the reason for the improved accuracy using standardized data for each subject is that, as mentioned in the hypothesis, when discriminating facial expressions from video, humans capture and judge relative changes of facial condition.
And We think the reason why the score improved by reducing the number of dimensions is thought to be that over-fitting occurred in the 512-dimensional time series data.

\begin{table*}
\caption{RESULT ON THE VALIDATION SET}
\label{table:validation1}
\begin{center}
\begin{tabular}{clllp{0pt}clll}
\hline\\[-6pt]
~ & \multicolumn{5}{c}{} & \multicolumn{3}{c}{EXPR} \\
\cline{7-9}\\[-6pt]
Method & \multicolumn{1}{c}{Multi-frame} & \multicolumn{1}{c}{Standardize} & \multicolumn{1}{c}{Audio} & \multicolumn{1}{c}{GRU size} & \multicolumn{1}{c}{Image dim} &
\multicolumn{1}{c}{\textbf{Score}} & \multicolumn{1}{c}{F1} & \multicolumn{1}{c}{Acc.}\\
\hline\hline
\\[-6pt]
Baseline \cite{cmp01} & - & - & - & - & - & 0.366 & 0.300 & 0.500 \\[-10pt]
TSAV \cite{abaw3} & - & - & - & - & - & 0.546 & - & - \\[-10pt]
Single-frame & - & - & - & 10 & 512 & 0.504 & 0.460 & 0.594 \\[2pt]
Multi-frame & \checkmark & - & - & 10 & 512 & 0.524 & 0.466 & 0.641 \\[2pt]
Multi-frame-std & \checkmark & \checkmark & - & 10 & 512 & 0.530 & 0.484 & 0.623 \\[2pt]
Multi-modal-std & \checkmark & \checkmark & \checkmark & 10 & 512 & 0.533  & 0.480  & 0.639 \\[2pt]
Multi-modal-std2 & \checkmark & \checkmark & \checkmark & 15 & 512 & 0.534  & 0.477  & 0.649 \\[2pt]
Multi-modal-std3 & \checkmark & \checkmark & \checkmark & 10 & 300 & 0.537  & 0.477  & 0.659 \\[2pt]
Multi-modal-std4 & \checkmark & \checkmark & \checkmark & 15 & 300 & \textbf{0.546}  & \textbf{0.488}  & \textbf{0.663} \\[2pt]
\hline
\end{tabular}
\end{center}
\end{table*}

\draftyamamoto{
The result of valence-arousal estimation using validation data is 0.245 and 0.442, respectively.
}

%%%%%%%%%%%%%%%%%%%%%%%%%%%%%%%%%%%%%%%%%%%%%%%%%%%%%%%%%%%%%%%%%%%%%%%%%%%%%%%%
\section{\large{C}\normalsize{ONCLUSIONS} AND \large{F}\normalsize{UTURE} \large{W}\normalsize{ORK}}

This paper describes the multi-modal analyzing framework for estimation of facial expression classifications using the Aff-Wild2 dataset. We introduced time-series data after combining the common features and the standardized features for each video into our framework. The verification results reveal that our proposed framework has achieved significantly higher performance than baseline on tracks 2 of the ABAW Challenge.
\\
In the future, we will consider applying other open source data sets and data expansion by using data with intentionally added occlusion to further improve accuracy and robustness.

%%%%%%%%%%%%%%%%%%%%%%%%%%%%%%%%%%%%%%%%%%%%%%%%%%%%%%%%%%%%%%%%%%%%%%%%%%%%%%%%

\nocite{*} % to test all bib entrys
%\bibliographystyle{unsrt} % <======================== not longer needed!
%\bibliography{\jobname} % <========================== not longer needed!
 % <======================================= mwe.bbl

\end{document}